# Large Language Models aren't all that you need

Kiran Voderhobli Holla
Indian Institute of Technology,
Hyderabad, India
CS21MDS14024@iith.ac.in

Chaithanya Kumar
Indian Institute of Technology,
Hyderabad, India
CS21MDS14022@iith.ac.in

Aryan Singh
Indian Institute of Technology,
Hyderabad, India
CS21MDS14029@iith.ac.in

## ABSTRACT

This paper describes the architecture and systems built towards solving the SemEval 2023 Task 2: MultiCoNER II (Multilingual Complex Named Entity Recognition) [1]. We evaluate two approaches (a) a traditional Conditional Random Fields model and (b) a Large Language Model (LLM) fine-tuned with a customized head and compare the two approaches. The novel ideas explored are: 1) Decaying auxiliary loss (with residual) - where we train the model on an auxiliary task of Coarse-Grained NER and include this task as a part of the loss function 2) Triplet token blending - where we explore ways of blending the embeddings of neighboring tokens in the final NER layer prior to prediction 3) Task-optimal heads - where we explore a variety of custom heads and learning rates for the final layer of the LLM. We also explore multiple LLMs including GPT-3 and experiment with a variety of dropout and other hyperparameter settings before arriving at our final model which achieves micro & macro f1 of **0.85/0.84** (on dev) and **0.67/0.61** on the test data . We show that while pre-trained LLMs, by themselves, bring about a large improvement in scores as compared to traditional models, we also demonstrate that tangible improvements to the Macro-F1 score can be made by augmenting the LLM with additional feature/loss/model engineering techniques described above.

## 1 Introduction

The Named Entity Recognition (NER) task aims to detect entities from unstructured text and classify them into predefined categories. Processing complex and ambiguous Named Entities (NEs) in low-context situations is a non-trivial NLP task. Malmasi et al., 2022b introduced the SemEval MultiCoNER shared task [2] that deals with multilingual complex NER in low context setting i.e., there are very few words in each sample. The SemEval 2023 MultiCoNER II competition compounds the complexity by using Fine-Grained labels instead of Coarse-Grained unlike the 2022 competition. Thus, complex named entities, like the titles of Creative Works are further broken down into fine-grained constituents like visual-work, art-work, musical-work etc., making the classification even more difficult with significant potential for noisy labels.

This paper describes our approach to tackle complex NER task for the English language using traditional as well as state of-the-art deep learning models. We also experiment with a variety of LLMs, a variety of custom heads coupled with a variety of loss functions. As of Feb 2023, any model having over 100 million parameters was considered large, so our definition of an LLM is any model that has over 100 million parameters. Our best scoring model has a BiLSTM + CRF head on top of an XLM-RoBERTa-Large LM which is trained at a learning rate of 10X as compared to the other layers. We use the token embeddings from the last 3 layers of the LLM instead of just the last layer. This is based on the original Bert paper (Devlin et al 2016) [3] where the authors show the merits of using information present in the last few penultimate layers of the LLM. During prediction, we blend the neighboring token information (Triplet token blending) so that the model has better context of the immediate neighbors. We also train parallelly on an auxiliary task of Coarse-Grained predictions and devise a custom loss function which is a combination of CRF loss for the original task and a decaying CRF loss (with residual) for the auxiliary task. We compare multiple architectures using the validation set of the shared task. We also run our final model(s) on the test dataset.

Most our transformer models outperform the baseline by a significant margin. Through our experiments, we discover that leveraging LLMs coupled with the above feature/model/loss engineering techniques result in excellent performance for fine-grained NER even in low-context settings without the need to augment with gazettes or external contexts. We run our models on a single seed/single fold and do not ensemble the results. Our model achieves micro-f1 scores of 0.43 & 0.67 for regular/transformer architectures respectively and 0.41 / 0.61 macro-f1 score on the final test dataset which by itself lands in the top 20 of the competition's CONLL 2023 leader board [4]. Based

Large Language models aren't all that you need on last year's competition analysis, we find that adding a context would have further boosted our scores by up to 0.1 which could have landed us in Top 10. Training with external data and creating an ensemble of models would significantly improve our scores even further.

We describe the prior research work in Section 2, formal task description in Section 3, the dataset details and our EDA in Section 4, the architecture in Section 5, experiments and ablations in Section 6, final results and error analysis in Sections 7 and finally, we conclude the paper in Section 9

## 2 PRIOR BODY OF WORK

NER is an essential tool for Information Retrieval, Question Answering (Banerjee et al., 2019) [5] and text summarization tasks. The traditional NER scheme is rule-recognition based on domain dictionary, grammar, heuristics, lexicons, orthographic features, and ontologies. Lample (2016) [6] explored bidirectional LSTMs combined with CRFs with features based on character-based word representations and unsupervised word representations. These do not require sources such as lexicons or ontologies, thus making them domain independent. Despite good results at that time, the computation was serial which hinders execution time.

Since 2019 LLMs like BERT (Devlin et al., 2019) [3] have shot into prominence and have been used to further boost the performance of NER, yielding state-of-the-art performances. The embeddings for LLMs are obtaining by pre-training on large-scale unlabeled data such as Wikipedia, which can significantly improve the contextual representations of named entities. Interestingly enough, the lack of context in the MultiCoNER I and II datasets mean that this particular advantage of LLMs is not fully exploited and hence, a host of different approaches were devised in last years' competition to add additional context to each sample. The winners of the MultiCoNER competition in 2022 - Damo-NLP [7] used an external context to augment each sample before making predictions. They do so by building a local Knowledge Base based on Wikipedia.

## 3 TASK DESCRIPTION: MultiCoNER II

The original CoNLL-2003 [8] language-independent NER was composed of 4 types of NEs. More recently, in WNUT 2017 [9], the task was expanded to identifying more novel and emerging entities. These types of entities are complex and depend on the surrounding context. Last year's SemEval 2022 MultiCoNER competition brings in the additional dimension of complexity of low resources. MultiCoNER II makes the task even more difficult by expanding the number of NE's to 36. The entities are fine-grained and could overlap with each other creating a potential for noisy labels and thus make the training more difficult.

For the monolingual track, the participants have to train a model that works for a single language. Train and dev sets are provided with labelled entities. The monolingual model trained needs to be used for the prediction of named entities in the test set. We have considered the English NER dataset for our task. Care is taken not to use the test dataset during training/validation phase and evaluation is done only on our final model(s) to benchmark our scores against the leader board.

## 4 DATASET DETAILS

MultiCoNER II provides uncased data from three domains (Wikipedia sentences, questions, and search queries) across 11 languages, which are used to define 11 monolingual subsets of the shared task. NER 2023 taxonomy allows us to capture a wide array of entities at a fine-grained level. BIO tagging scheme is followed. MultiCoNER II contains 36 fine-grained classes as opposed to MultiCoNER (2022) which contained just 6.

- The statistics for the English dataset are:
    - Train: 16,778; Dev: 871; Test: 249,980 sentences
- Accounting for B and I tags, the number of labels are 36 * 2 + one more for the Non-Entity ("O") leading to 73 classes. The English NER dataset (train + valid) contains 67 of these.
- We note that the training sample is low context and most samples are less than 50 words with a vast majority of sentences in the 10-20 range (Fig 1)

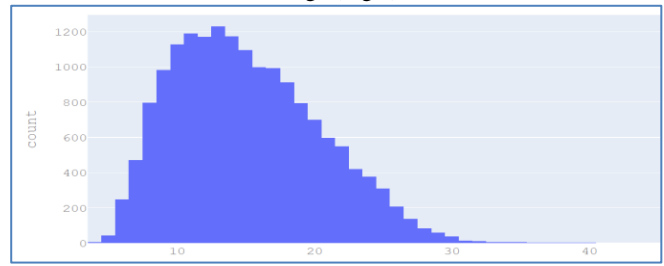

**Figure 1: Sentence size distribution in the given training data**

- We note that couple of sub-tags from PER (in particular 'artist') and many sub-tags from CW(Creative Works) dominate the number of tags (Fig 2)

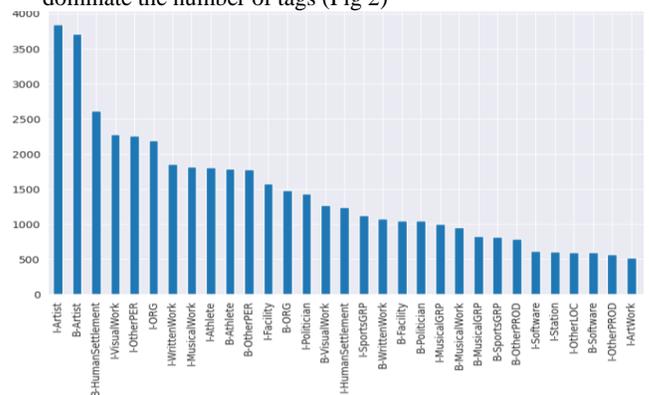

**Figure 2: Most frequently occurring tags in the training data**

- The distribution is long-tailed with many tags having very few mentions.

Large Language models aren't all that you need

- The test dataset is 15 times the size of train. The validation set provided is only around 5% the size of train. The model needs to be able to generalize well and perform without overfitting on the limited data available.

## 5 MODEL ARCHITECTURE

We model this task as a multiclass classification problem. The goal is for the model to predict the one amongst 67 final classes. We incorporate 2 main approaches – (a) Classic model (b) Large Language models (LLM).

### 5.1 Classic Model

We first manually create features for each sample. Some of the features we have included are (a) the previous word (b) whether the previous word is a digit, the Part Of Speech (POS) tag of the previous word etc. (c) Last few characters of the word and similar features for the previous word as well so better context can be built. All these features along with the original sample is then fed to the CRF model. We then use *sklearn_crfsuite.CRF* library to fit and predict the results.

CRF Lafferty et al. (2001) [10] is a statistical model which considers the neighboring samples by modelling the prediction as a graphical model. It assumes that the tag for the present word is dependent on the tag of its previous/next word. This model yields a MacroF1 score of 0.41. We contrast this traditional model with the LLM model in section 7.

### 5.2 Large Language Model

In this approach we leverage the embeddings of different language models as a starting point. The key workflow is as follows:

1. Use the embeddings of a pre-trained language model. We start with RoBERTa-base and substitute the language model with progressively larger ones retaining the rest of the architecture as-is and report our observations.
2. The pre-trained LLM is then fine-tuned for the task of NER using a custom head along with a host of additional feature and loss engineering. We expand this step in detail later in this section.
3. The final layer of the custom head is a simple linear layer where the number of output classes = 67. Alternatively, we also experiment with a CRF as a final layer.

The LLM with the highest score was XLM-RoBERTa-L. It was pre-trained on 2.5TB of filtered CommonCrawl data containing 100 languages by Facebook AI team. The pre-training was done using Masked language modeling (MLM) objective wherein the model randomly masks 15% of the words in the input then run the entire masked sentence through the model and has to predict the masked words. This pre-trained model was then released for fine-tuning on downward tasks. All the models we experimented with are obtained from the hugging face-transformers library which provides a standard interface to access these models.

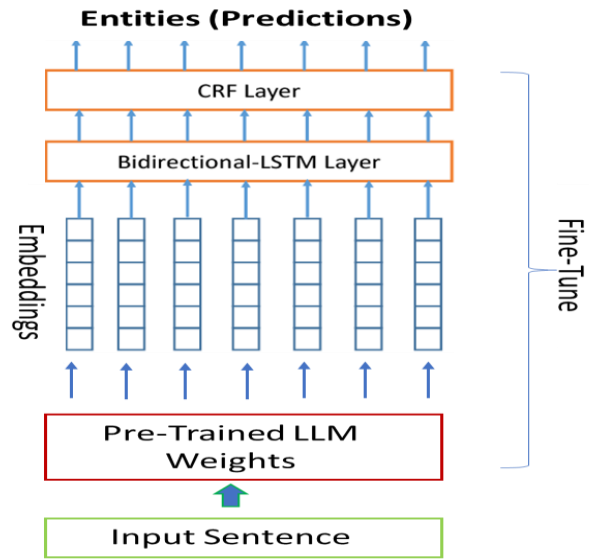

Figure 3: LLM(Large Language Model) Architecture

The following ideas are explored as part of the model architecture shown in Figure 3 above.

*5.2.1 Decaying auxiliary loss (with residual)*
Here we train the model on an auxiliary task of Coarse-Grained(CG) NER. We prepare a set of 6 CG labels and update the label for each sample. We now have 2 labels against each sample – CG and FG (Fine Grained). We then train the model on both the main task of FG predictions as well as the auxiliary task of CG predictions. The loss function is modified to include gradients for both tasks. Initially we start by giving 100% weightage to the CG task and none for the FG task but as we progress in terms of epochs, we alter this balance. We don't decay CG task weightage to 0 but retain an optimal residual.

$$Loss = W * (CG\ Loss) + (1 - W) * Scale * (FG\ Loss)$$

where W decays from 1 to a residual of 0.1 as the epochs progress and Scale is needed since the magnitude of the 2 terms vary due to the tag sizes.

We take this approach since the training data is less and labels are high. The model may not be able to generalize well in the absence of adequate training data for all labels. In such cases, the auxiliary task of CG NER will help the model learn more about the domain. For e.g., the embeddings generated with a CG aux task would ensure that musical-work is closer to art-work (as both fall under Creative Work CG category) and both are far off from (say) a PER-Politician. In a pure FG training, the model treats all 3 labels as different with no relation between all 3.

*5.2.2 Triplet token blending*

Large Language models aren't all that you need

Here we explore ways of blending the embeddings of neighboring tokens in the final NER layer prior to prediction. In conventional NER architectures, the last layer is a linear layer and the predictions for a token is based on the penultimate layer embeddings for that token. Since the penultimate token embeddings come from the transformer, which has self-attention layers built-in, the model is already aware of neighboring tokens to a certain degree. However, we felt there is a need to pay extra-special attention to the immediately preceding and succeeding neighbors during prediction and hence feed the information from these neighboring tokens to the model's final layer in addition to the original token embedding. We call this triplet token blending and use two approaches to blend – Concatenation and averaging. We find that concatenation works better empirically.

*5.2.3 Task-optimal heads*
Here, we explore a variety of custom heads and learning rates (LR) for the final layer of the large language model. The combinations explored are:

- Simple linear layer with masked CE loss
- Linear layer with masked CE loss + CRF layer
- Bi-LSTM + CRF layer

All 3 combinations above are trained at (a) Same LR (b) A higher LR (10X) for Head. Higher LR has been found to benefit because the head is completely untrained whereas the LLM is already pretrained. This helps reduce chances of catastrophic forgetting.

We also explore the option of picking the LLM embeddings from the last layer versus picking the embeddings from the last few penultimate layers of the LLM and concatenating them. We assumed that the later approach helps in better generalization.

# 6  EXPERIMENTS AND ABLATIONS

We conducted more than two dozen experiments and a (shortened) summary of the results are presented below. We analyze these results further in the next section.

| Config | LLM | LLM EMB used | Custom HEAD | Aux Task (CG) | Loss | Sep LR for Head |
|---|---|---|---|---|---|---|
| 1 | Roberta Base | Last layer | Linear | No | (Masked) Cross Entropy | No |
| 2 | XLM-Roberta-Large | Last layer | Linear | No | (Masked) Cross Entropy | No |
| 3 | XLM-Roberta-Large | Last layer | CRF | No | CRF loss | No |
| 4 | XLM-Roberta-Large | Last layer | CRF | No | CRF loss | 10 times PLM |
| 5 | Deberta-v3-large | Last layer | CRF | No | CRF loss | 100 times PLM |
| 6 | XLM-Roberta-Large | Last 3 layers | BiLSTM + CRF | No | CRF loss | 10 times PLM |
| 7 | XLM-Roberta-Large | Last layer | Linear | Linear | (Masked) CE + decaying CE Aux loss(with residual) | No |
| 8 | XLM-Roberta-Large | Last layer | Linear with Triplet blend (concat) | Linear | (Masked) CE + decaying CE Aux loss(with residual) | No |
| 9 | XLM-Roberta-Large | Last layer | Linear with Triplet blend (avg) | Linear | (Masked) CE + decaying CE Aux loss(with residual) | No |
| 10 | GPT-3 | NA | NA | NA | NA | NA |
| 11 | **XLM-Roberta-Large** | **Last 3** | **Bi-LSTM + CRF with Triplet blend (concat)** | **CRF** | **CRF loss + decaying CRF Aux loss (with residual)** | **10 times PLM** |

The scores for each of the above 'Config' is outlined below.

| Config | BS/EPOCHS/ DROPOUT | Dev Micro F1 | Dev Macro F1 | Test Micro F1 | Test Macro F1 |
|---|---|---|---|---|---|
| 1 | 32/5/0.1 | 0.53 | 0.29 | 0.54 | 0.31 |
| 2 | 32/5/0.1 | 0.69 | 0.57 | 0.65 | 0.53 |
| 3 | 32/5/0.1 | 0.7 | 0.6 | 0.66 | 0.55 |
| 4 | 16/10/0.1 | **0.86** | **0.84** | **0.67** | 0.59 |
| 5 | 16/10/0.1 | 0.75 | 0.67 | 0.58 | 0.48 |
| 6 | 16/10/0.1 | 0.85 | 0.83 | 0.66 | 0.59 |
| 7 | 16/10/0.1 | 0.84 | 0.8 | 0.66 | 0.58 |
| 8 | 16/10/0.1 | 0.84 | 0.8 | 0.66 | 0.58 |
| 9 | 16/10/0.1 | 0.84 | 0.8 | 0.66 | 0.58 |
| 10 | NA | 0.31 | 0.3 | NA* | NA* |
| 11 | 16/20/0.2 | **0.85** | **0.84** | **0.67** | **0.61** |

* Test data run on GPT-3 could not be completed. However, we present the (Dev) results to compare GPT-3 model with other LLMs as far as Dev metrics are concerned.

# 7  FINAL RESULTS AND ANALYSIS

Our best scoring model had the following scores on the test dataset:

- micro@F1    0.67
- micro@P     0.68
- micro@R     0.67
- macro@F1    0.61
- MD@R        0.85 (MD refers to the NER span predictions)
- MD@P        0.86
- MD@F1       0.86

We infer the following observations primarily based on the Macro F1 score which is used by the sponsors of MultiCoNER II to rank the leader board.

Large Language models aren't all that you need

1. LLM models beat traditional models by a fair margin. We believe this is due to extensive pre-training information and contextual embeddings which are available in LLMs which help improve the scores
2. The advantage of traditional model is that it carries a smaller memory and execution footprint and doesn't need GPU.
3. Traditional models need more feature engineering to reach good scores. The scores of our best performing traditional model are only slightly better than a very basic LLM with minimal feature engineering (RoBERTa-base). However, the scores jump significantly upward as we increase the size of the LM.
4. RoBERTa base gives a score of 0.31 which increases to 0.53 by simply replacing the LM with XLM-RoBERTa-Large.
5. It must however be noted that feature engineering, model engineering and loss engineering do play a role in further increasing the score from 0.53 to 0.61 which accounts for the title of our paper.
6. Having a higher dropout helps generalize better which is an important requirement of MultiCoNER II dataset considering the low-context data and low train data size.
7. Training the head with a faster LR helps improve the scores marginally. We found that the LR can be 10 times that of the LR of the LLM for optimal scores. LLMs are pretrained and don't need a very high learning rate.
8. CRF layer consistently improves the scores by a small factor due to the potential to capture and leverage transition data.
9. The auxiliary loss helps improve the Macro F1 score while retaining the micro F1 score as-is.
10. Since this is a low-context competition and most sentences are 50 tokens or lesser, a Bi-LSTM layer helps build additional attention mechanisms across the final layer embeddings (of the whole sentence) and makes a marginal improvement to the score. Bi-LSTM effect may get diluted when sentence lengths are large as information starts to leak.
11. Picking the embeddings of last 3 layers as opposed to picking the embeddings of the last layer do not help improve the scores. However, we believe that when building a multi-lingual system which is trained on far more data in different languages, the penultimate layer information could start making a difference to the scores.
12. We didn't see a noticeable difference in scores by leveraging the neighborhood blending option. This could be because the CRF layer handles all aspects of the transition mechanisms between two neighboring tokens already. However, we feel this is an area which can be explored further.
13. Counter-intuitively, we found that Creative Works (CW) F1 score is consistently on the higher side for nearly all sub-tags for LLM architecture. We attribute this to the fact discovered in EDA that there are higher number of tags associated with these in the training set and the LLM model is able to understand the context better compared to traditional model.
14. HumanSettlement, Artist tags have amongst the highest F1 score which could be a combination of the higher training data available coupled by the fact that these are unambiguous during predictions.
15. Interestingly, we also found nearly every sub-tag of PER NE with lower F1 (except Artist). While predictions for PER are expected to be unambiguous, we believe the lack of training data is hampering the scores. Adding a context would significantly boost the scores of such tags.
16. Though DeBERTa V3 L is the state-of-the-art LLM, we were unable to obtain a good score. DeBERTa V3 is trained on a different objective of Replaced Token Detection (RTD) instead of Masked Language Modelling, but we don't believe this should hamper the fine tuning on NER. We leave further investigations on this to a later day.
17. GPT-3 inference takes a significant amount of time and may be not be the most efficient choice for large volumes of data.

## 8 CONCLUSION

In this paper, we conclude that LLMs set a strong baseline for NER tasks even in fine-grained, low context scenarios and account for a huge uptick in scores. We also demonstrate that feature engineering, model engineering and loss engineering techniques continue to play an important role in further improving the scores. In particular, we show that building domain knowledge into the model by adding an auxiliary task of CG tag identification helps improve the Macro F1 scores.